\documentclass{article} 
\usepackage{nips13submit_e,times}
\usepackage{hyperref}
\usepackage{url}  
\usepackage{comment}  
\usepackage{subcaption}  

\newcommand{\exEnd}{$\Box$} 
\newtheorem{theorem}{Theorem} 

\newtheorem{definition}{Definition}

\newcounter{Examplecount}
\newenvironment{example}[1][Example]
{
\begin{trivlist}
\vspace{-1.6mm}
\refstepcounter{Examplecount}
\item[\hskip \labelsep{\bfseries #1}]{\small \textbf {\arabic{Examplecount}}. }}
{\end{trivlist} \vspace{-1.5mm}}
\usepackage{amssymb}
 \usepackage{comment}
\usepackage{xcolor}
\usepackage{graphicx}
\usepackage{float}
\newfloat{operator}{th}{lop}
\floatname{operator}{Operator}

\newfloat{algorithm}{t}{lop}
\floatname{algorithm}{Algorithm}

\title{First-Order Decomposition Trees}
\author{Nima Taghipour  \qquad  Jesse Davis  \qquad  Hendrik Blockeel\\
Department of Computer Science, KU Leuven\\
Celestijnenlaan 200A, B-3001 Heverlee, Belgium
}

\nipsfinalcopy

\begin{document}
\maketitle

\begin{abstract}
Lifting attempts to speedup probabilistic inference by exploiting symmetries in the model. Exact lifted inference methods, like their propositional counterparts, work by recursively decomposing the model and the problem. In the propositional case, there exist formal structures, such as decomposition trees (dtrees), that represent such a decomposition and allow us to determine the complexity of inference a priori. However, there is currently no equivalent structure nor analogous complexity results for lifted inference. In this paper, we introduce FO-dtrees, which upgrade propositional dtrees to the first-order level. We show how these trees can characterize a lifted inference solution for a probabilistic logical model (in terms of a sequence of lifted operations), and make a theoretical analysis of the complexity of lifted inference in terms of the novel notion of \emph{lifted width} for the tree. 
\end{abstract}

\section{Introduction}
\label{sec:intro}
Probabilistic logical modes (PLMs) combine elements of first-order logic with graphical models to succinctly model complex, uncertain, structured domains~\cite{Getoor07:book}. These domains often involve a large number of objects, making efficient inference a challenge. To address this, Poole~\cite{Poole2003} introduced the concept of {\em lifted probabilistic inference}, i.e., inference that exploits the symmetries in the model to improve efficiency. 
Various lifted algorithms have been proposed, mainly by \emph{lifting} propositional inference algorithms~\cite{Braz2005IJCAI,Gogate2011,Jha2010,kersting09uai,Milch2008,Poole2011,Singla2008,Taghipour2012,Taghipour2013b,GuyNips11,GuyIJCAI11,Venugopal12}. While the relation between the propositional algorithms is well studied, we have far less insight into their lifted counterparts. 

The performance of propositional inference, such as variable elimination~\cite{Dechter99,PooleZ03} or recursive conditioning~\cite{Darwiche01}, is characterized in terms of a corresponding {\em tree decomposition} of the model, 
and their complexity is measured based on properties of the decomposition, mainly its {\em width}. 
It is known that standard (propositional) inference has complexity exponential in the treewidth~\cite{Darwiche01,Dechter99}. 
This allows us to measure the complexity of various inference algorithms only based on the structure of the model and its given decomposition. Such analysis is typically done using a secondary structure for representing the decomposition of graphical models, such as decomposition trees ({\em dtrees})~\cite{Darwiche01}.  

However, the existing notion of treewidth does not provide a tight upper bound for the complexity of lifted inference, since it ignores the opportunities that lifting exploits to improve efficiency. Currently, there exists no notion analogous to treewidth for lifted inference to analyze inference complexity based on the model structure. 
In this paper, we take a step towards filling these gaps. 

Our work centers around a new structure for specifying and analyzing a lifted solution to an inference problem, and makes the following contributions. First, building on the existing structure of dtrees for propositional graphical models, we propose the structure of First-Order dtrees (FO-dtrees) for PLMs. An FO-dtree represents both the decomposition of a PLM and the symmetries that lifting exploits for performing inference. Second, we show how to determine whether an FO-dtree has a lifted solution, from its structure alone. Third, we present a method to read a lifted solution (a sequence of lifted inference operations) from a liftable FO-dtree, just like we can read a propositional inference solution from a dtree. Fourth,  we show how the structure of an FO-dtree determines the complexity of inference using its corresponding solution. We formally analyze the complexity of lifted inference in terms of the novel, symmetry-aware notion of {\em lifted width} for FO-dtrees. As such, FO-dtrees serve as the first formal tool for finding, evaluating, and choosing among lifted solutions.\footnote{Similarly to existing studies on propositional inference~\cite{Darwiche01,Dechter99}, 
our analysis only considers the 
model's {\em global structure}, and makes no assumptions about 
its {\em local structure}.}

\section{Background}
We use the term ``variable'' in both the logical and probabilistic sense. We use {\em logvar} for logical variables and {\em randvar} for random variables. We write variables in uppercase and their values in lowercase. Applying a substitution $\theta = \{s_1 \rightarrow t_1, \ldots, s_n \rightarrow t_n\}$ to a structure $S$ means replacing each occurrence of $s_i$ in $S$ by the corresponding $t_i$.  The result is written $S\theta$.

\subsection{Propositional and first-order graphical models}

Probabilistic graphical models such as Bayesian networks, Markov networks and factor graphs compactly represent a joint distribution over a set of randvars ${\mathcal V} = \{V_{1}, \ldots, V_{n}\}$ by factorizing the distribution into a set of local distribution. For example, factor graphs represent the distribution as a product of \emph{factors}: 
$\textit{Pr}(V_{1}, \ldots, V_{n}) = \frac{1}{Z} \prod \phi_i({\mathcal V_i})$, 
where $\phi_i$ is a \emph{potential} function that maps each configuration of ${\mathcal V}_i \subseteq \mathcal{V}$ to a real number and $Z$ is a normalization constant. 

Probabilistic logical models use concepts from first-order logic to provide a high-level modeling language for representing propositional graphical models. While many such languages exist (see~\cite{Getoor07:book} for an overview), we focus on \emph{parametric factors} (parfactors)~\cite{Poole2003} that generalize factor graphs. 

Parfactors use \emph{parametrized randvars} (PRVs) to represent entire sets of randvars. For example, the PRV $BloodType(X)$, where $X$ is a logvar, represents one $BloodType$ randvar for each object in the domain of $X$ (written $\mathcal{D}(X)$).   
Formally, a PRV is of the form $P({\mathbf X}) |  C$ where $C$ is a \emph{constraint} consisting of a conjunction of inequalities $X_i \neq t$ where $t \in \mathcal{D}(X_i)$ or $t \in {\mathbf X}$. 
It represents the set of all randvars $P({\mathbf x})$ where ${\mathbf x}\in \mathcal{D}({\mathbf X})$ and ${\mathbf x}$ satisfies $C$; this set is 
denoted $rv(P({\mathbf X})|C)$.

A {\em parfactor} uses PRVs to compactly encode a set of factors.  For example, the parfactor $\phi({\mathit Smoke}(X), {\mathit Friends}(X,Y),{\mathit Smoke}(Y))$ could encode that friends have similar smoking habits. 
It imposes a symmetry in the model by stating that the probability that, among two friends, both, one or none smoke, is the same for all pairs of friends, in the absence of any other information. 

Formally, a parfactor is of the form $\phi({\mathcal A})|C$, where ${\mathcal A} = (A_i)_{i=1}^n$ is a sequence of PRVs, $C$ is a constraint on the logvars appearing in $\mathcal{A}$, and $\phi$ is a potential function. The set of logvars occurring in ${\mathcal A}$ is denoted $logvar({\mathcal A})$. 
A {\em grounding substitution} maps each logvar to an object from its domain.
A parfactor $g$ represents the set of all factors that can be obtained by applying a grounding substitution to $g$ that is consistent with $C$; this set is called the grounding of $g$, and is denoted $gr(g)$. A parfactor model is a set $G$ of parfactors. It compactly defines a factor graph $gr(G) = \{gr(g) | g \in G\}$.

Following the literature, we assume that the model is in a \emph{normal form}, such that (i) each pair of logvars have either identical or disjoint domains, and 
(ii) for each pair of co-domain logvars $X$, $X'$ in a parfactor $\phi({\mathcal A}) |C$, $(X \neq X') \in C$. 
Every model can be written into this form in poly time~\cite{Poole2011}. 

\subsection{Inference}
A typical inference task is to compute the
marginal probability of some variables 
by summing out the remaining variables, which can be written as: 
$\textit{Pr}({\mathcal V}') = \sum_{{\mathcal V} \setminus {\mathcal V}'} \prod_i \phi_i({\mathcal V}_i)$. 
This is an instance of the general sum-product problem~\cite{Bacchus09}. Abusing notation, we write this sum of products as $\sum_{{\mathcal V}\setminus \mathcal{V}'} M(\mathcal{V})$. 

\noindent\textbf{Inference by recursive decomposition.} 
Inference algorithms exploit the factorization of the model to recursively decompose the original problem into smaller, independent subproblems. 
This is achieved by a decomposition of the sum-product, according to a simple {\em decomposition rule}. 

\begin{definition}[{\bf The decomposition rule}] Let ${\mathcal P}$ be a sum-product computation ${\mathcal P}: \sum_{{\mathcal V}} M({\mathcal V})$, 
and let $\mathbb{M} = \{M_1({\mathcal V_1}), \dots  M_k({\mathcal V_k})\}$ be a partitioning (decomposition) of $M({\mathcal V})$. 
Then, the {\em decomposition of ${\mathcal P}$, w.r.t.\ ${\mathbb M}$} is an equivalent sum-product formula ${\mathcal P}_{{\mathbb M}}$, defined as follows:
$${\mathcal P}_{{\mathbb M}}: \sum_{{\mathcal V}'} \Big[  \, \big( \sum_{{\mathcal V}'_1} M_1({\mathcal V}_1) \big) \dots  \big(\sum_{{\mathcal V}'_k} M_k({\mathcal V}_k) \big) \,  \Big]$$\vspace{-0.2cm}
where ${\mathcal V}' = \bigcup_{i,j} {\mathcal V}_i \cap {\mathcal V}_j$, and ${\mathcal V}'_i = {\mathcal V_i} \setminus {\mathcal V}'$.
\end{definition}

Most exact inference algorithms recursively apply this rule 
and compute the final result using top-down or bottom-up dynamic programming 
\cite{Bacchus09,Darwiche01,Dechter99}. The complexity is then exponential only in the size of the largest sub-problem solved. 
Variable elimination (VE) is a bottom-up algorithm that computes the nested sum-product by repeatedly solving an innermost problem $\sum_V M(V,\mathcal{V}')$ to \emph{eliminate} $V$ from the model. At each step, VE eliminates a randvar $V$ from the model by \emph{multiplying} the factors in $M(V,\mathcal{V}')$ into one and \emph{summing-out} $V$ from the resulting factor. 

\begin{figure} [t]
\centering
\includegraphics[height = 3cm]{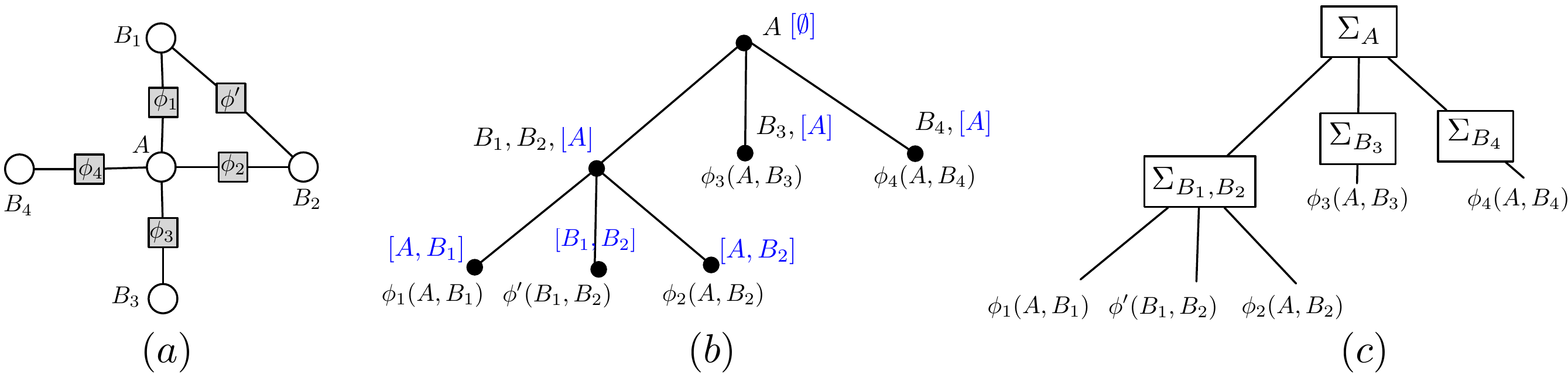}
\caption{(a) a factor graph model; (b) a dtree for the model, with its node clusters shown as $cutset, [context]$; (c) the corresponding factorization of the sum-product computations.}\vspace{-0.2cm}
\label{fig:dtree_nb}
\end{figure}

\noindent\textbf{Decomposition trees.} A single inference problem typically has multiple solutions, each with a different complexity. 
A \emph{decomposition tree} (dtree) is a structure that represents the decomposition used by a specific solution and allows us to determine its complexity~\cite{Darwiche01}.  Formally, a dtree is a rooted tree in which each leaf represents a factor in the model.\footnote{We use a slightly modified definition for dtrees, which were originally defined as full binary rooted trees.} 
 Each node in the tree represents a decomposition of the model into the models under its child subtrees. Properties of the nodes can be used to determine the complexity of inference. $Child(T)$ refers to $T$'s child nodes; 
$rv(T)$ refers to the randvars under $T$, which are those in its factor if $T$ is a leaf and $rv(T) = \cup_{T' \in Child(T)} rv(T')$ otherwise. 
Using these, the important properties of {\em cutset}, {\em context}, and {\em cluster} are defined as follows: 
\begin{itemize}
\item $cutset(T) = \cup_{\{T_1, T_2\} \in child(T)} rv(T_1) \cap rv(T_2) \setminus acutset(T)$, where ${\textit acutset}(T)$ is the union of cutsets associated with ancestors of $T$. \item  $context(T) = rv(T) \cap {\textit acutset}(T)$
\item $cluster(T) = rv(T)$, if $T$ is a leaf; otherwise $cluster(T) = cutset(T) \cup context(T)$
\end{itemize}
Figure~\ref{fig:dtree_nb} shows a factor graph model, a dtree for it with its clusters, and the corresponding sum-product factorization. Intuitively, the properties of dtree nodes help us analyze the size 
of subproblems solved during inference. In short, the time complexity of inference is $O(n \exp(w))$ where $n$ is the size (number of nodes) of the tree and $w$ is its \emph{width}, i.e., its maximal cluster size minus one.

\section{Lifted inference: Exploiting symmetries}
The inference approach of Section 2.2 ignores the symmetries imposed by a PLM. 
Lifted inference aims at exploiting symmetries among a model's {\em isomorphic} parts. Two constructs are isomorphic if there is a structure preserving bijection between their components. 
As PLMs make assertions about whole groups of objects, they contain 
many isomorphisms, established by a bijection at the level of objects. Building on this, symmetries arise between constructs at different levels~\cite{Niepert12}, such as between: randvars, value assignments to randvars, factors, models, or even sum-product problems. All exact lifted inference methods use two main tools for exploiting symmetries, i.e., for lifting: 
\begin{enumerate}
\item Divide the problem into isomorphic subproblems, solve one instance, and aggregate
\item Count the number of isomorphic configurations for a group of interchangeable variables instead of enumerating all possible configurations.
\end{enumerate}
\indent 
Below, we show how these tools are used by lifted variable elimination (LVE)~\cite{Braz2005IJCAI,Milch2008,Poole2003,Taghipour2012,Taghipour2013b}. 

\noindent\textbf{Isomorphic decomposition: exploiting symmetry among subproblems.} 
The first lifting tool identifies cases where the application of the decomposition rule results in a product of isomorphic sum-product problems. Since such problems all have isomorphic answers, we can solve \emph{one } problem and reuse its result for all the others. In LVE, this corresponds to lifted elimination, which uses the operations of \emph{lifted multiplication} and \emph{lifted sum-out} on parfactors to evaluate a single representative problem. Afterwards, LVE also attempts to aggregate the result (compute their product) by taking advantage of their isomorphism. For instance, when the results are identical, LVE computes their product simply by \emph{exponentiating} the result of one problem. 

\begin{figure}[t]
\begin{center}
\includegraphics[height = 1.3cm ]{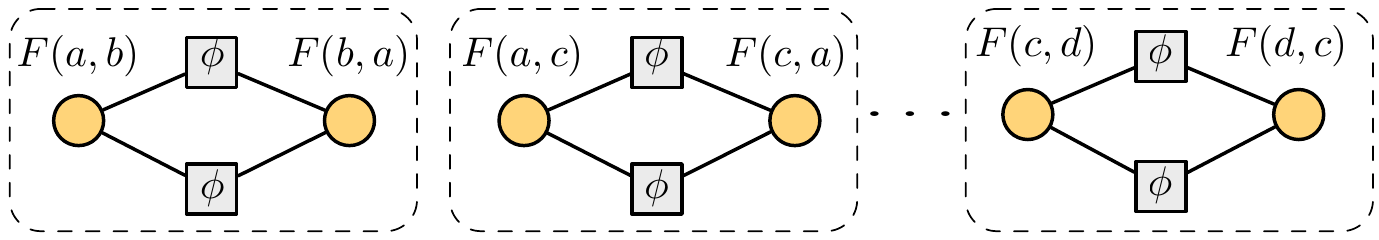}
\end{center}
\caption{Isomorphic decomposition of a model. Dashed boxes indicate the partitioning into groups.} \label{fig:group-inv}
\end{figure}

\begin{example}
\label{ex:group-inv}
Figure~\ref{fig:group-inv} shows the model defined by $\phi(F(X,Y), F(Y,X)) | X \neq Y$, with $\mathcal{D}(X)=\mathcal{D}(Y)=\{a,b,c,d\}$. The model asserts that the friendship relationship ($F$) is likely to be symmetric. 
To sum-out the randvars $F$ using the decomposition rule, we partition the ground factors into six groups of the form $\{\phi(F(x,y), F(y,x)), \phi(F(y,x), F(x,y))\}$, i.e., one group for each 2-subset $\{x,y\} \subseteq \mathcal \{a,b,c,d\}$. Since no randvars are shared between the groups, 
this decomposes the problem into the product of six isomorphic sums $\sum_{F(x,y),F(y,x)} \phi(F(x,y), F(y,x)) \cdot \phi(F(y,x), F(x,y))$. All six sums have the same result $c$ (a scalar). Thus, LVE computes $c$ only once (lifted elimination) and computes the final result by exponentiation as $c^6$ (lifted aggregation). 
\end{example}

\noindent\textbf{Counting: exploiting interchangeability among randvars.}
Whereas isomorphic decomposition exploits symmetry among problems, counting exploits symmetries within a problem, by identifying \emph{interchangeable} randvars. A group of ($k$-tuples of) randvars are {\em interchangeable}, if permuting the assignment of values to the group results in an equivalent model. 
Consider a sum-product subproblem $\sum_{\mathcal{V}} M(\mathcal{V},\mathcal{V}')$ that contains a set of $n$ interchangeable ($k$-tuples of) randvars ${\mathcal V} = \{(V_{i1}, V_{i2}, \dots V_{ik}) \}_{i=1}^n$. The interchangeability allows us to rewrite $\mathcal{V}$ into a single {\em counting randvar} $\#[{\mathcal V}]$, whose value is the histogram $h= \{({\mathbf v}_1, n_1), \dots, ({\mathbf v}_{r}, n_r)\}$, where $n_i$ is the number of tuples with joint state $\mathbf{v}_i$. This allows us to replace a sum over all possible joint states of $\mathcal{V}$ with a sum over the histograms for $\#[\mathcal{V}]$. That is, we compute $M({\mathcal V}')  = \sum_{i=1}^{m} \textsc{Mul}(h_i) \times  M(h_i, {\mathcal V}')$, where $\textsc{Mul}(h_i)$ denotes the number of assignments to ${\mathcal V}$ that yield the same histogram $h_i$ for $\#[{\mathcal V}]$. 
Since the number of histograms is $O(n^{\exp(k)})$, when $n \gg k$, we gain exponential savings over enumerating all the possible joint assignments, whose number is $O(\exp(n^{k}))$. This lifting tool is employed in LVE by \emph{counting conversion}, which rewrites the model in terms of counting randvars. 

\begin{example}
Consider the model defined by the parfactor $\phi(S(X),S(Y)) | X \neq Y$, which is $\prod_{i \neq j} \phi(S(x_i), S(x_j))$. 
The group of randvars 
$\{S(x_1), \dots, S(x_n)\}$ are interchangeable here, since under any value assignment where $n_t$ randvars are $true$ and $n_f$ randvars are $false$, the model evaluates to the same value 
$\phi'(n_t,n_f) = \phi(t,t)^{n_t.(n_t-1)} \cdot \phi(t,f)^{n_t. n_f} \cdot \phi(f,t)^{n_f.n_t} \cdot \phi(f,f)^{n_f.(n_f-1)}$. By counting conversion, LVE rewrites this model into $\phi'(\#_X[S(X)])$. 
\end{example}

\section{First-Order decomposition trees}
In this section, we propose the structure of FO-dtrees, which compactly represent a recursive decomposition for a PLM and the symmetries therein.

\subsection{Structure}
An FO-dtree provides a compact representation of a propositional dtree, just like a PLM is a compact representation of a propositional model. It does so by explicitly capturing isomorphic decomposition, which in a dtree correspond to a node with isomorphic children. Using a novel node type, called a \emph{decomposition into partial groundings (DPG)} node, an FO-dtree represents the \emph{entire set} of isomorphic child subtrees with a \emph{single representative subtree}. To formally introduce the structure, we first show how a PLM can be decomposed into isomorphic parts by DPG. 

\noindent\textbf{DPG of a parfactor model.} 
The DPG of a parfactor $g$ is defined w.r.t.\ a $k$-subset ${\mathbf X} = \{X_1, \dots, X_k \}$ of its logvars that all have the same domain $D_{\bf X}$. 
For example, the decomposition used in Example~\ref{ex:group-inv}, and shown in Figure~\ref{fig:group-inv}, is the DPG of $\phi(F(X,Y),F(Y,X))|X\neq Y$ w.r.t.\ logvars $\{X,Y\}$. 
Formally, $DPG(g,{\bf X})$ partitions the model defined by $g$ into ${|D_{\bf X}| \choose k}$ parts: one part $G_{{\bf x}}$ for each $k$-subset ${\bf x} = \{x_1, \dots, x_k\}$ of the objects in $D_{\bf X}$. Each $G_{\bf x}$ in turn contains all $k!$ (partial) groundings of $g$ that can result from replacing $(X_1, \dots, X_k)$ with a permutation of $(x_1, \dots, x_k)$. 
The key intuition behind DPG is that  for any ${\bf x},{\bf x}' \subseteq_k \mathcal{D}_{\bf X}$, $G_{{\bf x}}$ is isomorphic to $G_{{\bf x}'}$, 
since any bijection from ${\bf x}$ to ${\bf x}'$ yields a bijection from $G_{{\bf x}}$ to $G_{{\bf x}'}$. 

$DPG$ can be applied to a whole model $G= \{ g_i\}_{i=1}^m$, 
if $G$'s logvars are (re-)named 
such that (i) only co-domain logvars share the same name, and (ii) logvars ${\mathbf X}$ appear in all parfactors. 

\begin{example} 
\label{ex:dpg2}
Consider $G = \{\phi_1(P(X) )$, $\phi_2(A,P(X))\}$. $DPG(G,\{X\}) = \{G_i\}_{i=1}^n$, where each group $G_i = \{\phi_1(P(x_i) )$, $\phi_2(A,P(x_i))\}$ is a grounding of $G$ (w.r.t.\ $X$). 
\end{example}

\begin{figure}[tb]
\centering
\includegraphics[height = 3 cm]{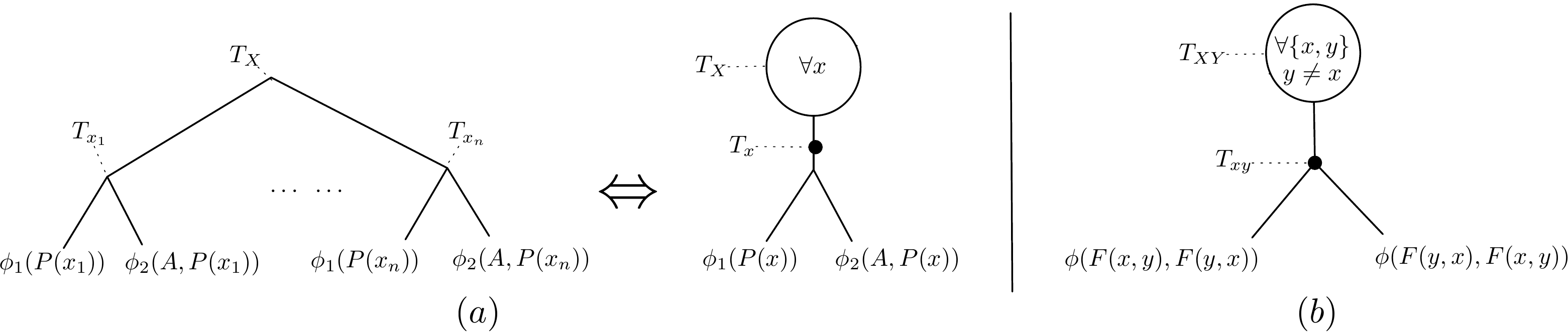}
\caption{(a) dtree (left) and FO-dtree (right) of Example~\ref{ex:dpg2}; (b) FO-dtree of Example~\ref{ex:group-inv}}
	\label{fig:simple_fodt} 
\end{figure}

FO-dtrees simply add to dtrees special nodes for representing DPGs in parfactor models. 

\begin{definition}[DPG node] A DPG node $T_{{\bf X}}$ is a triplet $({\mathbf X}, {\mathbf x}, C)$, where ${\bf X} = \{X_1, \dots X_k \}$ is a set of logvars with the same domain $D_{\mathbf X}$, ${\mathbf x} = \{x_1, \dots, x_k \}$ is a set of {\em representative objects}, 
and $C$ is a constraint, such that for all $i\neq j$: $x_i \neq x_j \in C$. We denote this node as $\forall {\bf x}:C$ in the tree.
\end{definition}

A representative object is simply a placeholder for a domain object.\footnote{As such, it plays the same role as a logvar. 
However, we use both 
to distinguish between a whole group of randvars (a PRV $P(X)$), and a representative of this group (a representative randvar $P(x)$).} 
The idea behind our FO-dtrees is to use $T_{{\bf X}}$ to graphically indicate a $DPG(G, {\bf X})$. 
For this, each $T_{\bf X}$ has a single child distinguished as $T_{{\bf x}}$, under which the model is a representative instance of the isomorphic models $G_{\bf x}$ in the DPG. 

\begin{definition}[FO-dtree] 
An FO-dtree is a rooted tree in which 
\begin{enumerate} 
\item non-leaf nodes may be DPG nodes
\item each leaf contains a factor (possibly with representative objects) 
\item each leaf with a representative object $x$ is the descendent of exactly one DPG node $T_{\bf X} = ({\mathbf X}, {\mathbf x}, C)$, such that $x \in {\bf x}$ 
\item each leaf that is a descendent of $T_{\bf X}$ has all the representative objects ${\bf x}$, and 
\item for each $T_{{\bf X}}$ with ${\bf X} = \{X_1, \dots, X_k \}$, $T_{\bf x}$ has $k!$ children $\{ T_i \}_{i=1}^{k!}$, which are isomorphic up to a permutation of the representative objects ${\bf x}$. 
\end{enumerate}
\end{definition}

\noindent{\bf Semantics.} Each FO-dtree defines a dtree, which can be constructed by recursively \emph{grounding} its DPG nodes. 
Grounding a DPG node $T_{{\bf X}}$ yields a (regular) node $T'_{{\bf X}}$ with ${|\mathcal{D}_{\bf X}| \choose k}$ 
children $\{T_{{\bf x} \rightarrow {\bf x}'}| {\bf x'} \subseteq_k D_{\bf X} \}$, where 
$T_{{\bf x} \rightarrow {\bf x}'}$ is the result of replacing ${\bf x}$ with objects ${\bf x'}$ in $T_{{\bf x}}$. 

\begin{example} Figure~\ref{fig:simple_fodt} (a) shows the dtree of Example~\ref{ex:dpg2} and its corresponding FO-dtree, which only has one instance $T_x$ of all isomorphic subtrees $T_{x_i}$. Figure~\ref{fig:simple_fodt} (b) shows the FO-dtree for Example~\ref{ex:group-inv}.
\end{example}

\subsection{Properties}
Darwiche~\cite{Darwiche01} showed that important properties of a recursive decomposition are captured in the properties of dtree nodes. 
In this section, we define these properties for FO-dtrees. 
Adapting the definitions of the dtree properties, such as cutset, context, and cluster, for FO-dtrees requires accounting for the semantics of
an FO-dtree, which uses DPG nodes and representative objects. More specifically, this requires making the following
two modifications (i) use a function $Child_{\theta}(T)$, instead of $Child(T)$, to take into account the semantics of DPG nodes, and (ii) use a function $\cap_{\theta}$ that finds the intersection of two sets of {\em representative} randvars. 
First, for a DPG node $T_{{\bf X}} = ({\bf X}, {\bf x}, C)$, 
we define: $Child_{\theta}(T_X) = 
\{ T_{{\bf x} \rightarrow {\bf x}'} | {\bf x}' \subseteq_k {\mathcal D}_{\bf X} \}$. 
Second, for two sets $A = \{a_i\}_{i=1}^n$ and $B= \{b_i\}_{i=1}^n$ of (representative) randvars we define:
$A \cap_{\theta} B = \{a_i  | \exists \theta \in \Theta: a_i \theta \in B\},$ 
with  $\Theta$ the set of grounding substitutions to their representative objects. Naturally, this provides a basis to define a `$\setminus_{\theta}$' operator as : $A \setminus_{\theta} B = A \setminus (A \cap_{\theta} B)$.

\begin{figure}[htb]
\centering
\includegraphics[height = 4 cm]{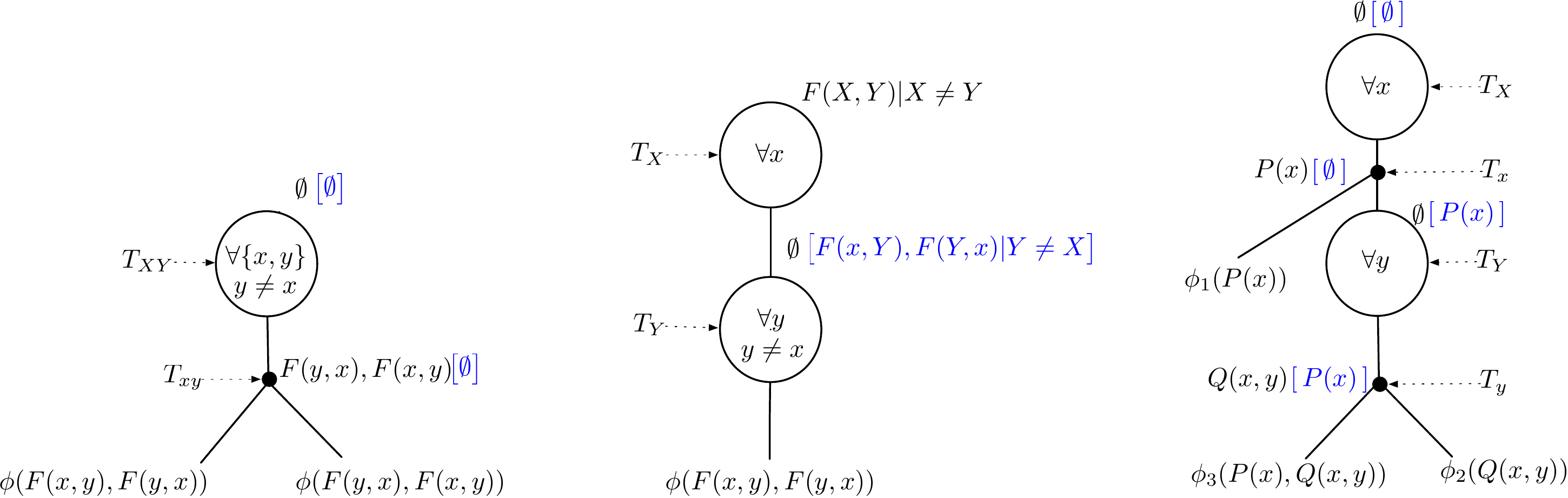}
\caption{Three FO-dtree with their clusters (shown as $cutset, [context]$).} 
\label{fig:2lv_fodt_props}
\end{figure}

All the properties of an FO-dtree are defined based on their corresponding definitions for dtrees, by replacing $Child$, $\cap$, $\setminus$ with $Child_{\theta}$, $\cap_{\theta}$, $\setminus_{\theta}$. 
Interestingly, all the properties can be computed without grounding the model, e.g., for a DPG node $T_X$, we can compute $rv(T_X)$ simply as $rv(T_x) \theta^{-1}_X$, with $\theta^{-1}_X = \{{\mathbf x} \rightarrow \mathbf{X}\}$.\footnote{The only non-trivial property is $cutset$ of DPG nodes. 
We can show that $cutset(T_X)$ excludes from $rv(T_X) \setminus acutset(T_X)$ only those PRVs for which ${\bf X}$ is a binding class of logvars~\cite{Jha2010,GuyNips11}.} 
Figure~\ref{fig:2lv_fodt_props} shows examples of FO-dtrees with their node clusters. 

\begin{figure}[hbt]
\centering
\includegraphics[height = 3.2 cm]{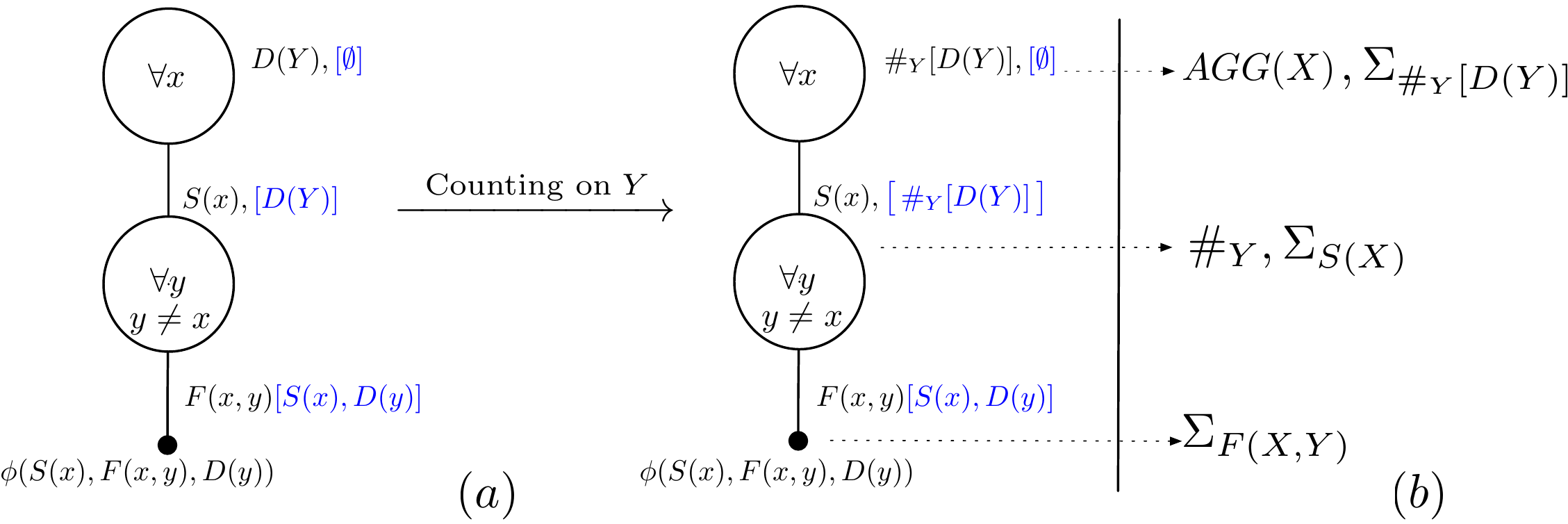}
\caption{(a) an FO-dtree (left) and its counted version (right); 
(b) lifted operations of each node.}
\label{fig:fodt_count_ops}
\end{figure}

\noindent\textbf{Counted FO-dtrees.} 
FO-dtrees capture the first lifting tool, isomorphic decomposition, explicitly in DPG nodes. The second tool, counting, 
can be simply captured by rewriting interchangeable randvars in clusters of the tree nodes with counting randvars. This can be done in FO-dtrees similarly to the operation of counting conversion on logvars in LVE. We call such a tree a {\em counted} FO-dtree. 
Figure~\ref{fig:fodt_count_ops}(a) shows an FO-dtree (left) and its counted version (right). 

\vspace{-0.4cm}
\section{Liftable FO-dtrees}
When inference can be performed using the lifted operations (i.e., without grounding the model), it runs in polynomial time in the domain size of logvars. Formally, this is called a domain-lifted inference solution~\cite{GuyNips11}. Not all FO-dtrees have a lifted solution, which is easy to see since not all models are liftable~\cite{Jaeger2012}, though each model has at least one FO-dtree.\footnote{A basic algorithm for constructing an FO-dtree for a PLM is presented in the appendix.} 
Fortunately, we can structurally identify the FO-dtrees for which we know a lifted solution. 

\noindent\textbf{What models can the lifting tools handle?} Lifted inference identifies isomorphic problems and solves only one instance of those. 
Similar to propositional inference, for a lifted method the difficulty of each sub-problem increases with the number of variables in the problem-- those that appear in the {\em clusters} of FO-dtree nodes. When each problem has a bounded (domain-independent) number of those, the complexity of inference is clearly independent of the domain size. However, a sub-problem can involve a large group of randvars--- when there is a PRV in the cluster. 
While traditional inference is then intractable, lifting may be able to exploit the interchangeability among the randvars and reduce the complexity by \emph{counting}. 
Thus, whether a problem has a lifted solution boils down to whether we can rewrite it such that it only contains a bounded (domain-independent) number of counting randvars and ground randvars. This requires the problem to have enough symmetries in it such that all the randvars $\mathcal{V} = V_1, \dots V_n$ in each cluster can be divided into $k$ groups of interchangeable (tuples of) randvars $\mathcal{V}_1, \mathcal{V}_2, \dots ,\mathcal{V}_k$, where $k$ is independent of the domain size. 

\begin{theorem} A (non-counted) FO-dtree has a lifted inference solution if its clusters only consist of (representative) randvars and $1$-logvar PRVs. We call such an FO-dtree a {\em liftable} tree.\footnote{Note that this only restricts the number of logvars in PRVs appearing in an \emph{FO-dtree's clusters}, not PRVs in the PLM. For instance, all the liftable trees in this paper correspond to PLMs containing 2-logvar PRVs. 
}
\end{theorem}

{\em Proof sketch.} Such a tree has a corresponding LVE solution: (i) each sub-problem that we need to solve in such a tree can be formulated as a (sum-out) problem on a model consisting of a parfactor with $1$-logvar PRVs, and (ii) we can count-convert all the logvars in a parfactor with $1$-logvar PRVs~\cite{Milch2008,Taghipour_StarAI12}, to rewrite all the PRVs into a (bounded) number of counting randvars.\footnote{For a more detailed proof, see the appendix.} 
\vspace{-0.2cm}

\section{Lifted inference based on FO-dtrees}
A dtree can prescribe the operations performed by propositional inference, such as VE~\cite{Darwiche01}. 
In this section, we show how a liftable FO-dtree can prescribe an LVE solution for the model, thus providing the first formal method for symbolic operation selection in lifted inference.

In VE, each inference procedure can be characterized based on its elimination order. 
Darwiche~\cite{Darwiche01} shows how we can read a (partial) elimination order 
from a dtree (by assigning elimination of each randvar to some tree node). 
We build on this result to read an LVE solution from a (non-counted) FO-dtree. For this, we assign to each node a set of lifted operations, including lifted elimination of PRVs (using multiplication and sum-out), and counting conversion and aggregation of logvars: 

\begin{itemize}
\item $\sum_\mathcal{V}$: A PRV $\mathcal{V}$ is \emph{eliminated} at anode $T$, if $\mathcal{V} \in cluster(T) \setminus context(T)$.
\item $AGG(X)$: A logvar $X$ is {\em aggregated} at a DPG node $T_{\bf X} = ({\bf X}, {\bf x}, C)$, if (i) $X \in {\bf X}$, and (ii) $X \notin logvar(cluster(T_{\bf X}))$.
\item $\#_X$: A logvar $X$ is {\em counted} at $T_{\bf X}$, if (i) $X \in {\bf X}$, and (ii) $X \in logvar(cluster(T_X))$.
\end{itemize}

A lifted solution can be characterized by a sequence of these operations. For this we simply need to order the operations according to two rules: 
\begin{enumerate}
\item If node $T_2$ is a descendent of $T_1$, and $OP_i$ is performed at $T_i$, then $OP_2 \prec OP_1$. 
\item For operations at the same node, aggregation and counting precede elimination. 
\end{enumerate}

\begin{example} From the FO-dtree shown in Figure~\ref{fig:fodt_count_ops} (a) 
we can read the following order of operations: 
$\sum F(X,Y) \prec \#_Y \prec \sum S(X) \prec AGG(X) \prec \sum \#_Y[D(Y)]$, see Figure~\ref{fig:fodt_count_ops} (b). 
\exEnd \end{example}

\section{Complexity of lifted inference}
In this section, we show how to compute the complexity of lifted inference based on an FO-dtree. 
Just as the complexity of ground inference for a dtree is parametrized in terms of the tree's {\em width}, we define a {\em lifted width} for FO-dtrees and use it to parametrize the complexity of lifted inference. 

To analyze the complexity, it suffices to compute the complexity of the operations performed at each node. Similar to standard inference, this depends on the randvars involved in the node's {\em cluster}: for each lifted operation at a node $T$, LVE manipulates a factor involving the randvars in $cluster(T)$, and thus has complexity proportional to $O(|\textit{range}(\textit{cluster}(T))|)$, where $\textit{range}$ denotes the set of possible (joint) values that the randvars can take on. However, unlike in standard inference, this complexity need not be exponential in $|rv(\textit{cluster}(T))|$, since the clusters can contain {\em counting randvars} that allow us to handle interchangeable randvars more efficiently. 
To accommodate this in our analysis, we 
define two {\em widths} for a cluster: a {\em ground width} $w_g$, which is the number of ground randvars in the cluster, and a {\em counting width}, $w_{\#}$, which is the number of counting randvars in it. The cornerstone of our analysis is that the complexity of an operation performed at node $T$ is exponential only in $w_g$, and polynomial in the domain size with degree $w_\#$. We can thus compute the complexity of the entire inference process, by considering the hardest of these operations, and the number of operations performed. 
We do so by defining a \emph{lifted width} for the tree. 

\begin{definition}[{\bf Lifted width}] The \emph{lifted width} of an FO-dtree $T$ is a pair $(w_g, w_{\#})$, where $w_g$ is the largest ground width among the clusters of $T$ and and $w_{\#}$ is the largest counting width among them.
\end{definition}
\vspace{-0.1cm}
\begin{theorem} The complexity of lifted variable elimination for a counted liftable FO-dtree $T$ is:
$$O(n_T \cdot \log n \cdot \exp(w_g) \cdot n_{\#} ^ {(w_{\#} \cdot r_{\#})}),$$
where $n_T$ is the number of nodes in $T$, $(w_g, w_{\#})$ is its lifted width, $n$ (resp., $n_{\#}$) is the 
the largest domain size among its logvars (resp., counted logvars), and $r_{\#}$ is the largest range size among its tuples of 
counted randvars. 
\end{theorem} 
\vspace{-0.1cm}
\noindent\emph{Proof sketch.}
We can prove the theorem by showing that (i) the largest range size among clusters, and thus the largest factor constructed by LVE, is $O(\exp(w_g) \cdot n ^ {(w_{\#} \cdot r_{\#})})$, (ii) in case of aggregation or counting conversion, each entry of the factor is exponentiated, with complexity $O(\log n)$, and (iii) there are at most $n_T$ operations.  (For a more detailed proof, see the appendix.)
\exEnd

\noindent\textbf{Comparison to ground inference.} 
To understand the savings achieved by lifting, it is useful to compare the above complexity to that of standard VE on the corresponding dtree, i.e., using the same decomposition. 
The complexity of ground VE is: 
$O(n_G \cdot  \exp(w_g) \cdot \exp(n_{\#} . w_{\#}))$, where $n_G$ is the size of the corresponding propositional dtree. 
Two important observations are: \vspace{-0.1cm}
\begin{enumerate}
\item The number of ground operations is linear in the dtree's size $n_G$, instead of the FO-dtree's size $n_T$ (which is polynomially smaller than $n_G$ due to DPG nodes). Roughly speaking, lifting allows us to perform $n_T/n_G$ of the ground operations by \emph{isomorphic decomposition}. 
\item Ground VE, has a factor $\exp(n_{\#} . w_{\#})$ in its complexity, instead of $n_{\#}^{w_{\#}}$ for lifted inference. The latter is typically exponentially smaller. These speedups, achieved by \emph{counting}, are the most significant for lifted inference, and what allows it to tackle high treewidth models. 
\end{enumerate}

\section{Conclusion}
We proposed FO-dtrees, a tool for representing a recursive decomposition of PLMs. An FO-dtree explicitly shows the symmetry between its isomorphic parts, and can thus show a form of decomposition that lifted inference methods employ. We showed how to decide whether an FO-dtree is liftable (has a corresponding lifted solution), and how to derive the sequence of lifted operations and the complexity of LVE based on such a tree. While we focused on LVE, our analysis is also applicable to lifted search-based methods, such as lifted recursive conditioning~\cite{Poole2011}, weighted first-order model counting~\cite{GuyIJCAI11}, and probabilistic theorem proving~\cite{Gogate2011}. This allows us to derive an order of operations and complexity results for these methods, when operating based on an FO-dtree. Further, we can show the close connection between LVE and search-based methods, by analyzing their performance based on the same FO-dtree. FO-dtrees are also useful to 
approximate lifted inference algorithms, such as lifted blocked Gibbs sampling~\cite{Venugopal12} and RCR~\cite{guyuai12}, 
that attempt to improve their inference accuracy by identifying liftable subproblems and handling them by exact inference. 

\section*{Appendix}
\appendix

In this appendix, we provide proofs for the Theorem 1 and 2, and present a basic algorithm for constructing FO-dtrees for PLMs. 

\section{Proof of Theorem 1}

\emph{Proof.} Following the discussion in the paper, each subproblem arising during inference requires handling a parfactor involving the randvars and PRVs that appear at the cluster of the node. To prove that each of these problems are liftable (do not require us to ground the PRVs and deal with all their randvars directly), we need to show that  the whole group of randvars in each cluster can be partitioned into $m$ groups of interchangeable $k$-tuples of randvars, with $m$ and $k$ independent of the domain size. 
We prove this relying on the properties of counting randvars in PLMs, and the correctness of counting conversion in LVE~\cite{Milch2008,Taghipour_StarAI12}. 
For simplicity, let us assume that there are no ground randvars in the cluster (the generalization to include ground randvars is trivial). 
Then the model can be written as a $1$-logvar parfactor as follows:
$$ \phi(P_{11}(X_{11}), \dots P_{1,n_1}(X_{1,n_1}), \dots,  P_{m1}(X_{11}),  \dots P_{m,n_1}(X_{m,n_m})) \, | \, C,$$
in which for each $i \in \{1,\dots,m\}$, all $X_{ij}$ are logvars from a distinct domain $D_i$, and $P_{ij}$ is an PRV containing such a logvar---note that for the same $i$ some $X_{ij}$ (and some $P_{ij}$) can have the same name, although the PRVs are distinct. 
Since no PRV contains more than one logvar we can count-convert all the logvars in this model. This merges all distinct PRVs $P_{ij}(X_i)$ into one counting randvar. As such, by applying counting conversion on all the logvars $X_{ij}$ of domain $D_i$, we can rewrite in the model the group of PRVs $P_{i1}(X_{i1}), \dots, P_{i,n_i}(X_{i,n_i})$ into a counting randvar $$\#_{X_i}[P'_{i1}(X_i), \dots, P'_{i,k_i}(X_i)]$$
where $P'_{ij}$ are the distinct predicates among $P_{ij}$, that is:
$$\{P'_{ij}(X_i)\}_{j=1}^{k_i} = \bigcup_{j=1}^{n_i} P_{ij}(X_i)$$

After counting all the logvars the parfactor becomes of the form 
$$\phi' \big( \#_{X_1}[P'_{11}(X_1), \dots, P'_{1,k_1}(X_1)], \, \dots \,, \#_{X_m}[P'_{m1}(X_m), \dots, P'_{m,k_m}(X_m)] \big)$$

This shows that the whole group of randvars in the model can be partitioned into $m$ groups of interchangeable $k$-tuples of randvars-- one group of tuples for each counting randvar. Note that here both $k$ and $m$ are independent of the domain size of the logvars: (i) $m$ is the number of distinct domains among the logvars, and (ii) $k$ can be no larger than the number of PRVs with a co-domain logvar in the model, that is, $k \leq \max \{k_i\}_i \leq \max \{n_i\}_i$. It is straight-forward to show that this also holds in the general case of a parfactor involving both $1$-logvar and ground randvars.

\section{Proof of Theorem 2}

\emph{Proof.} We prove the theorem by bounding the complexity of each lifted operation performed at each of the $n_T$ nodes of the tree. First consider a lifted elimination performed at  some node $T'$. The complexity of this operation is proportional to $|\mathit{range}(cluster(T'))|$, as it needs to deal with a parfactor involving the (counting) randvars in the cluster. Each cluster is a group $\mathcal{A} = \{A_1, A_2, \dots A_{w_{g}'}, \gamma_1, \gamma_2, \dots, \gamma_{w_{\#}'} \}$ of randvars $A_i$, and counting randvars $\gamma_i = \#_{X_i}[P_{i1}(X_i), \dots, P_{ik}(X_{i})]$, where $w_{\#}' \leq w_{\#}$, and $w_{g}' \leq w_{g}$.
Thus $$|range(\mathcal{A})| = \big(\prod_i |range(A_i)|\big) \cdot \big( \prod_j |range(\gamma_j)|\big).$$
 
For the first product, we have $$\prod_{i=1}^{w_{g}'} |range(A_i)| = O(\exp(w_{g})).$$ 
Moreover,  since for each counting randvar $\gamma_i$, $|range(\gamma_i)| = O(n_i^{r_i})$, where $n_i$ is the domain size of $X_i$, and  $r_i$ is the range size of the tuples of PRVs inside $\gamma_i$, for the second product we have 
$$\prod_{j=1}^{w_{\#}'} |range(\gamma_j)| = O((n_{\#}^{r_{\#}})^{w_{\#}}) = O(n_{\#} ^ {(w_{\#} \cdot r_{\#})})$$
These two show that  
$$|\mathit{range}(\mathcal{A})| = O(\exp(w_{g}) \cdot n_{\#} ^ {(w_{\#} \cdot r_{\#})})$$

This is the complexity of each lifted elimination step. 
Build on this we compute the complexity of the other two lifted operations, aggregation and counting conversion. For each of the $|\mathit{range}(\mathcal{A})|$ entries in the parfactor, these two operations perform an exponentiation which has complexity $O(\log n)$, where $n$ is the domain size of the logvar. As such, this has complexity $O(\log n \cdot \exp(w_g) \cdot n_{\#} ^ {(w_{\#} \cdot r_{\#})})$. Since there at most one of each operation performed at each of the $n_T$ nodes, the complexity of entire inference is
$$O(n_T \cdot \log n \cdot \exp(w_g) \cdot n_{\#} ^ {(w_{\#} \cdot r_{\#})}).$$ 

\section{Finding corresponding FO-dtrees}

In this section, we provide a simple algorithm that given a model $G$ constructs a corresponding FO-dtree. Our method works in a top-down manner according to a recursive decomposition of $G$ using $DPGs$. We also briefly discuss possible extensions of this simple algorithm, which can transform it into a greedy algorithm for finding `better' trees. 

We construct the tree top-down according to a recursive decomposition of $G$, which also employs $DPGs$ (Algorithm~\ref{alg:fodt_simple}). At the beginning we have a single root node $T$ with model $G$. According to a decomposition of $G$ into $\{G_i\}_i$ we add the children $T_i$ of $T$ to the tree, and then recursively build each tree $T_i$ for $G_i$. Under DPG nodes we represent only one instance of the children. DPGs allow us to decompose the model into partial groundings, and recursive application of this tool results in a ground model. This allows us to reduce the problem to finding a dtree for the ground model. 

\begin{algorithm}[htb]
\begin{center}
\begin{tabular}{l}
\hline
{\bf FO-dtree}($G$)\\
{\bf if} $G$ is ground\\ 
\quad {\bf return} \textsc{Dtree}($G$)\\
{\bf if} $\exists {\bf X}$ that allows DPG\\
\quad $T_{\bf X} \leftarrow \textsc{DPG-Node}({\mathbf X}, {\mathbf x}, G)$\\
\quad $G_{\mathbf x} = \{G \theta | \theta \in \Theta_{\mathbf{x}}\}$\\
\quad $T.\textsc{addChild}$({\bf FO-Dtree}($G_{\mathbf x}))$\\
{\bf else}:\\ 
\quad $T \leftarrow \textsc{newnode}()$\\
\quad choose logvars {\bf X} that co-occur in $G$:\\
\quad (there is always at least one choice ${\bf X} = \{X_i\}$)\\
\quad $G_{{\bf X}} \leftarrow \{g | {\bf X} \in logvar(g)\}$\\ 
\quad $G_{\neg {\bf X}} \leftarrow G \setminus G_{{\bf X}}$\\
\quad $T.\textsc{addChildren}(${\bf FO-Dtree}$(G_{\mathbf X}), ${\bf FO-Dtree}$(G_{\neg \mathbf X}))$\\
{\bf return} $T$\\
\hline
\end{tabular}
\end{center}
\caption{A simple algorithm for finding a corresponding FO-dtree.}
\label{alg:fodt_simple}
\end{algorithm} 

\textbf{Extension to a greedy method for finding FO-dtrees.} The above is a simple algorithm that shows the existence of a FO-dtree for each model, by finding one possible FO-dtree. While it does not consider the quality of the found FO-dtree, it can be easily modified into an algorithm that greedily searches for better trees, by performing better DPGs. For this we need to make two changes in Algorithm~\ref{alg:fodt_simple}: (1) rename the logvars such that the model allows for a DPG, instead of relying on the naming of logvars in the model, and (2) select among the possible DPGs based on some criteria. 

The first change requires us to {\em align} the logvars in different parfactors before performing a DPG, that is to rename the logvars properly such that a subset of the logvars allow for DPG. This is a simple generalization of finding an alignment between two parfactors, which is employed in lifted multiplication. This change allows us to consider all possible DPGs of the model in our search, without being restricted by the naming of logvars in the model. 
The second change allows us to consider the quality of different DPGs for selection among them. Here we give a score to possible DPGs, which is a greedy measure of the quality of their decomposition. For instance, we can simply consider the cutset size of the decomposition, or the size of its resulting clusters. A straightforward measure is comparing the lifted width of the resulting nodes, which takes into account also the opportunities exploited by counting. 
These two changes should be naturally incorporated into one module, which considers possible logvar re-namings (alignments) that enable some DPG, measures the quality of the corresponding DPGs, and selects among them. Search for alignments can be guided by considering the properties of logvars in the model~\cite{Jha2010,GuyNips11}, and our result about computing properties of FO-dtree nodes based on the properties of logvars.

\bibliographystyle{plain}
\bibliography{LiftedInference}

\begin{thebibliography}{10}

\bibitem{Bacchus09}
F.~Bacchus, S.~Dalmao, and T.~Pitassi.
\newblock Solving \#-{SAT} and {B}ayesian inference with backtracking search.
\newblock {\em Journal of Artificial Intelligence Research}, 34(2):391, 2009.

\bibitem{Darwiche01}
Adnan Darwiche.
\newblock Recursive conditioning.
\newblock {\em Artif. Intell.}, 126(1-2):5--41, 2001.

\bibitem{Braz2005IJCAI}
Rodrigo {de Salvo Braz}, Eyal Amir, and Dan Roth.
\newblock Lifted first-order probabilistic inference.
\newblock In {\em Proceedings of the 19th International Joint Conference on
  Artificial Intelligence (IJCAI)}, pages 1319--1325, 2005.

\bibitem{Dechter99}
Rina Dechter.
\newblock Bucket elimination: {A} unifying framework for reasoning.
\newblock {\em Artif. Intell.}, 113(1-2):41--85, 1999.

\bibitem{Getoor07:book}
Lise Getoor and Ben Taskar, editors.
\newblock {\em An Introduction to Statistical Relational Learning}.
\newblock {MIT} {P}ress, 2007.

\bibitem{Gogate2011}
Vibhav Gogate and Pedro Domingos.
\newblock Probabilistic theorem proving.
\newblock In {\em Proceedings of the 27th Conference on Uncertainty in
  Artificial Intelligence (UAI)}, pages 256--265, 2011.

\bibitem{Jaeger2012}
Manfred Jaeger and Guy {Van den Broeck}.
\newblock Liftability of probabilistic inference: {U}pper and lower bounds.
\newblock In {\em Proceedings of the 2nd International Workshop on Statistical
  Relational AI (StaRAI)}, 2012.

\bibitem{Jha2010}
Abhay Jha, Vibhav Gogate, Alexandra Meliou, and Dan Suciu.
\newblock Lifted inference seen from the other side : {T}he tractable features.
\newblock In {\em Proceedings of the 23rd Annual Conference on Neural
  Information Processing Systems (NIPS)}, pages 973--981. 2010.

\bibitem{kersting09uai}
Kristian Kersting, Babak Ahmadi, and Sriraam Natarajan.
\newblock Counting belief propagation.
\newblock In {\em Proceedings of the 25th Conference on Uncertainty in
  Artificial Intelligence (UAI)}, pages 277--284, 2009.

\bibitem{Milch2008}
Brian Milch, Luke~S. Zettlemoyer, Kristian Kersting, Michael Haimes, and
  Leslie~Pack Kaelbling.
\newblock Lifted probabilistic inference with counting formulas.
\newblock In {\em Proceedings of the 23rd AAAI Conference on Artificial
  Intelligence (AAAI)}, pages 1062--1608, 2008.

\bibitem{Niepert12}
Mathias Niepert.
\newblock Markov chains on orbits of permutation groups.
\newblock In {\em Proceedings of the 28th Conference on Uncertainty in
  Artificial Intelligence (UAI)}, pages 624--633, 2012.

\bibitem{Poole2003}
David Poole.
\newblock First-order probabilistic inference.
\newblock In {\em Proceedings of the 18th International Joint Conference on
  Artificial Intelligence (IJCAI)}, pages 985--991, 2003.

\bibitem{Poole2011}
David Poole, Fahiem Bacchus, and Jacek Kisynski.
\newblock Towards completely lifted search-based probabilistic inference.
\newblock {\em CoRR}, abs/1107.4035, 2011.

\bibitem{PooleZ03}
David Poole and Nevin~Lianwen Zhang.
\newblock Exploiting contextual independence in probabilistic inference.
\newblock {\em J. Artif. Intell. Res. (JAIR)}, 18:263--313, 2003.

\bibitem{Singla2008}
Parag Singla and Pedro Domingos.
\newblock Lifted first-order belief propagation.
\newblock In {\em Proceedings of the 23rd AAAI Conference on Artificial
  Intelligence (AAAI)}, pages 1094--1099, 2008.

\bibitem{Taghipour_StarAI12}
Nima Taghipour and Jesse Davis.
\newblock Generalized counting for lifted variable elimination.
\newblock In {\em Proceedings of the 2nd International Workshop on Statistical
  Relational AI (StaRAI)}, 2012.

\bibitem{Taghipour2012}
Nima Taghipour, Daan Fierens, Jesse Davis, and Hendrik Blockeel.
\newblock Lifted variable elimination with arbitrary constraints.
\newblock In {\em Proceedings of the 15th International Conference on
  Artificial Intelligence and Statistics (AISTATS)}, pages 1194--1202, 2012.

\bibitem{Taghipour2013b}
Nima Taghipour, Daan Fierens, Guy {Van den Broeck}, Jesse Davis, and Hendrik
  Blockeel.
\newblock Completeness results for lifted variable elimination.
\newblock In {\em Proceedings of the 16th International Conference on
  Artificial Intelligence and Statistics (AISTATS)}, 2013.

\bibitem{GuyNips11}
Guy {Van den Broeck}.
\newblock On the completeness of first-order knowledge compilation for lifted
  probabilistic inference.
\newblock In {\em Proceedings of the 24th Annual Conference on Advances in
  Neural Information Processing Systems (NIPS)}, pages 1386--1394, 2011.

\bibitem{guyuai12}
Guy {Van den Broeck}, Arthur Choi, and Adnan Darwiche.
\newblock Lifted relax, compensate and then recover: From approximate to exact
  lifted probabilistic inference.
\newblock In {\em Proceedings of the 28th Conference on Uncertainty in
  Artificial Intelligence (UAI)}, pages 131--141, 2012.

\bibitem{GuyIJCAI11}
Guy {Van den Broeck}, Nima Taghipour, Wannes Meert, Jesse Davis, and Luc
  De~Raedt.
\newblock Lifted probabilistic inference by first-order knowledge compilation.
\newblock In {\em Proceedings of the 22nd International Joint Conference on
  Artificial Intelligence (IJCAI)}, pages 2178--2185, 2011.

\bibitem{Venugopal12}
Deepak Venugopal and Vibhav Gogate.
\newblock On lifting the gibbs sampling algorithm.
\newblock In {\em Proceedings of the 26th Annual Conference on Advances in
  Neural Information Processing Systems (NIPS)}, pages 1--6, 2012.

\end{thebibliography}

\end{document}